\documentclass[letterpaper, 11pt]{article}

\usepackage[utf8]{inputenc}
\usepackage[TS1,T2A, T1]{fontenc}

\usepackage[all]{xy}

\usepackage[english]{babel}
\usepackage[unicode=true]{hyperref}
\usepackage{amsmath}
\usepackage{amssymb}
\usepackage{amsthm}
\usepackage{graphicx}
\usepackage{url}
\usepackage{makecell}
\usepackage{algpseudocode}
\usepackage{cite}

\newtheorem{thm}{Special Theorem}
\theoremstyle{definition}
\newtheorem{defn}[thm]{Definition}
\newtheorem{reqr}[thm]{Requirement}

\newcommand*\xor{\mathbin{\oplus}}
\newcommand*\band{\mathbin{\&}}
\newcommand{\ie}{\textit{i.e.}}
\newcommand{\first}{\textbf} 
\newcommand{\term}{\textit}

\ifthenelse{\isundefined{\keywords}}{
	\def\keywords{\vspace{.5em}
	{\textit{Keywords}:\,\relax%
	}}
	
}{
}

\newcommand{\approptoinn}[2]{\mathrel{\vcenter{
  \offinterlineskip\halign{\hfil$##$\cr
    #1\propto\cr\noalign{\kern2pt}#1\sim\cr\noalign{\kern-2pt}}}}}

\newcommand{\appropto}{\mathpalette\approptoinn\relax}

\begin{document}
\title{Multilayered Model of Speech}
\date{}
\author{Andrey Chistyakov\footnote{\textit{Email}: andrey.chistyakov@ghoort.com}}
\maketitle

\begin{abstract}%
Human speech is the most important part of General Artificial Intelligence and subject of much research. The hypothesis proposed in this article provides explanation of difficulties that modern science tackles in the field of human brain simulation. The hypothesis is based on the author's conviction that the brain of any given person has different ability to process and store information. Therefore, the approaches that are currently used to create General Artificial Intelligence have to be altered.
\end{abstract}


\section{Introduction}
\label{sec:introduction}
The problem of human speech modeling has been solved more than once, but each time it faced paradoxes and contradictions. The scientific approach was initiated by Boolean algebra, which transgressed into logic of predicates. The human speech is more complicated than the predicate logic, but at the same time it is free of paradoxes, which occur at the first or higher orders. This article is an attempt to solve certain problems by introducing special restrictions.    

Speech analysis will be done in some consecutive stages. We should create automata, which can work over context--sensitive grammer, at first. This grammer will be defined in section~\ref{sec:basic_definitions}. This definition based on concept of mask, which is responsible to symbol from ingress aphabet. The automata's transition function devides on some parts, which has specific logic operations over its. The main difference with previous studies is the specific function $\Phi$, which can represent any class as data and some data as the class. The section~\ref{sec:layers_working} describe the automata's algorithm. There introduced the simple context as the set of symbols between two terminals.  

We describe the hierarchical system of one-modifying automatas in section~\ref{sec:context_working}. The automata system should recognise symbols in context-sensitive grammer. Analogies for the hierarchical system exists in scientific literature, but the article has the significant difference. The difference is in memory system with selfmodifications by the operations over the classes from section~\ref{sec:layers_working}. The selfmodification ability has some special restrictions because we need to pass over infinite looping of any logical algorithm, related to G\"{o}del theorem.
 
We now turn to generalized grammar in speech, generally independent of a specific language. Basic elements of speech are lexemes, they are the words with a meaning. We will use following lexemes - Nouns, Verbs and Adjectives. Nouns and Verbs are clear, because they exist in all human languages. Nouns names objects and subjects, Verbs - actions between them. Adjectives marks qualities and quantities. Lexemes do not exist by themselves, but only being bound to sentence with subjects, predicate, and other words, including function words. 

Adjectives represent some qualities of the object, which a person initially receives from feeling an interaction with an object, so it is possible to say that adjectives are linked with feelings. Verbs represent actions, which are performed by a person or any other subject. Those actions are usually performed through muscle reflexes. Therefore Verbs are linked with muscle reflexes. Nouns serve as markers of feelings/reflexes triggers. It will be shown from which they originate. Basic sentence are some stable sequences of ingress and egress signals. In this sense, words or lexemes, as letters sequences, are also lower level sentences. 

But speech consists of not only egress and ingress sentences, it also includes the human memory. Memory is presumed to be multilevel, objectual, and context oriented. It contains complex, processed reflexes. Memory is arranged in a hierarchical graph, its nodes are classes and objects. Class is a Noun collection of Adjectives and Verbs. Object is an implementation of class. Also classes can be represented as data and they may be included in the object. Based on this feature, classes can be converted between each other and packed into memory.

A concept of option is introduced for definition of memory work. This is the main innovation of the concept of the proposed model. In the option's scope there will be two possible options of information decryption. Both of these concepts are not equivalent and they can exist in two different people, but not together in the one person. Every option is fixed randomly, during the creation of the brain, and will not change during the whole lifetime of the person.

A total of four independent options will be introduced, which means that a possibility of 16 different options of human speech are possible in this model. It is the most significant difference from all other attempts of creating human speech modeling mechanisms. It is presumed  that this division was made possible in the result of evolution, for the interpretation of all incoming information, while every individual can intake only a part from the "raw" flow, to process and transfer to other individuals, prepared in way which can be used.

The possibility of introducing optionalization is grounded on different psychological theories, based on Karl Jung's archetypes~\cite{Jung}, primarily Myers--Briggs Type Indicator~\cite{MayerBriggs}. 

For the modeling of the brain I will use terms and principles from Computer Science, such as addressed memory, search operations of certain symbols in sequence and Boolean operations. Names for some terms are taken from Socionics, a special discipline about typology of the human personality.

\section{Basic Definitions}
\label{sec:basic_definitions}
Lets introduce basic definitions for the beginning.
\begin{defn}
Let a point be a pair from an address, where the address is a natural number, and value equal to $0$ or $1$.
\end{defn}
\begin{defn}
Layer is one-dimensional finite array of points, with consistent and continuous numbering of addresses from $0$ to $N$, where $N$ is the array length. The array length is fixed during the working process, but values of points are variable. 
\end{defn}
\begin{defn}
Let the stack be finite, consistent, and continuous array of layers.
\end{defn}
\begin{defn}
Mask $\mu$ is a fixed, finite array of a pair $(i,[0,1])$, where $i$ are natural numbers from $0$ to $N$.  
\end{defn}
It is necessary to clarify that even though the mask is consistent, \ie{} it has no points, which have equal addresses, but all addresses of the mask cannot be continuous. Mask's addresses should have gaps, and their meaning will be discussed below. 

Basic operations, in the system of layers and masks, are offset by definition and by blocking the mask range. Suppose that at some point of time values, at a layer, are unchangeable. Then it becomes possible to check all layers, and find a combination of values in the layer, which are represented in the mask. The least address element, of first point of the layer, appropriates to the mask, is called \first{begin of block} or \first{offset}. The last address is called as \first{end of block}. The gap between the beginning and the end - including, is called \first{block of layer}. A block of layers always matches to a specific mask. Operation of finding entry of a mask - is called obtaining an offset.
\begin{figure}[h]
\centering
\includegraphics[width=0.5\textwidth]{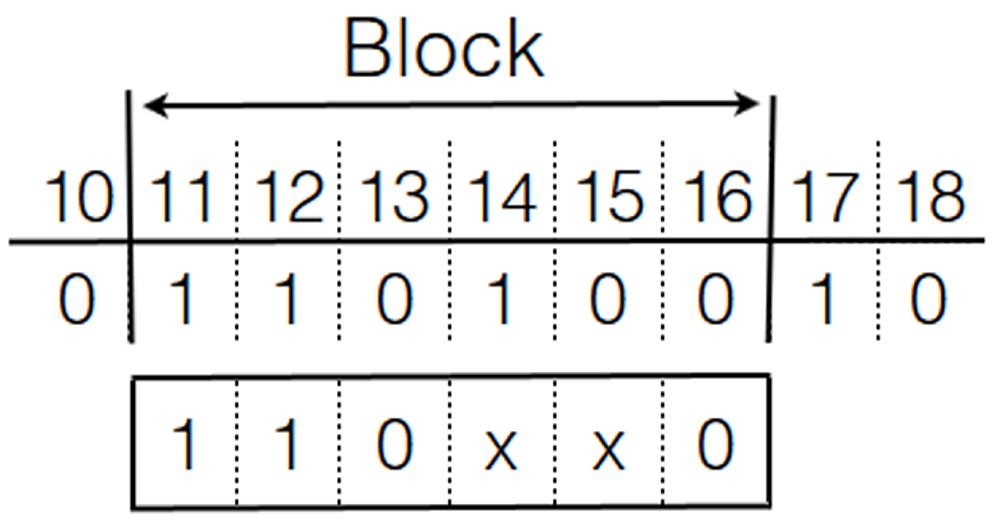}
\caption{Mask}
\end{figure}
\begin{reqr}
\label{reqr:non_intersect}
Blocks on the same layer should not intersect each other.
\end{reqr}
\begin{defn}
Let $B$ be a block of a layer, which consists of serial, continuous set $(b_0,b_1,\ldots,b_m)$. Then it is possible to define the set of the Boolean function $\mathfrak{F}(B)$ of the type $f(b_i,\ldots,b_k)\to[0,1]$. The set $\mathfrak{F}(B)$ will be closed under Boolean operations $\xor$ (XOR) and $\band$ (AND). Let's call $\mathfrak{F}(B)$ as set of \first{Adjectives} and the result of these values as \first{quantities}.   
\end{defn}
\begin{figure}[h]
\centering
\includegraphics[width=0.3\textwidth]{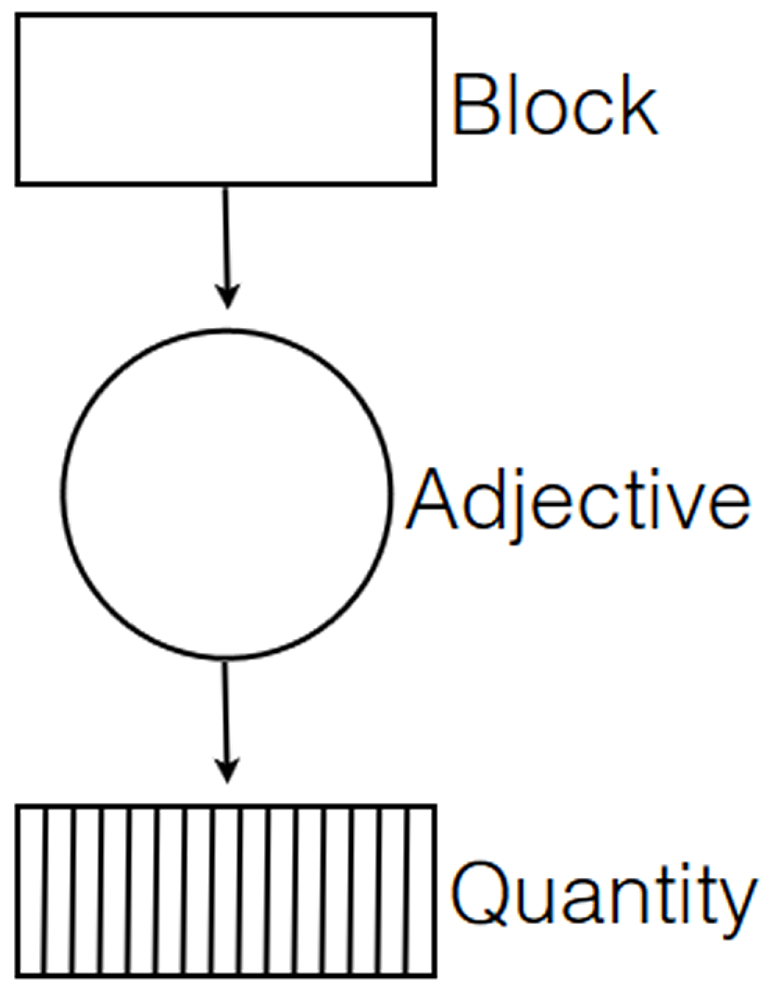}
\caption{Quality and Adjective}
\end{figure}
\begin{defn}
Let there be a set of qualities $Q = (q_0,\ldots\,q_n)$ and block $B$. Then it is possible to define the set of the Boolean function $\mathfrak{P}(Q,B)$ of the type $f(q_0,\ldots\,q_n,b_i,\ldots,b_k)\to[0,1]$. The set $\mathfrak{P}(Q,B)$ will be closed under the Boolean operations $\xor$ (XOR) and $\band$ (AND). Let's call $\mathfrak{P}(Q,B)$ as set of \first{Verbs} and the resulting values as \first{actions}.
\end{defn}
\begin{figure}[h]
\centering
\includegraphics[width=0.5\textwidth]{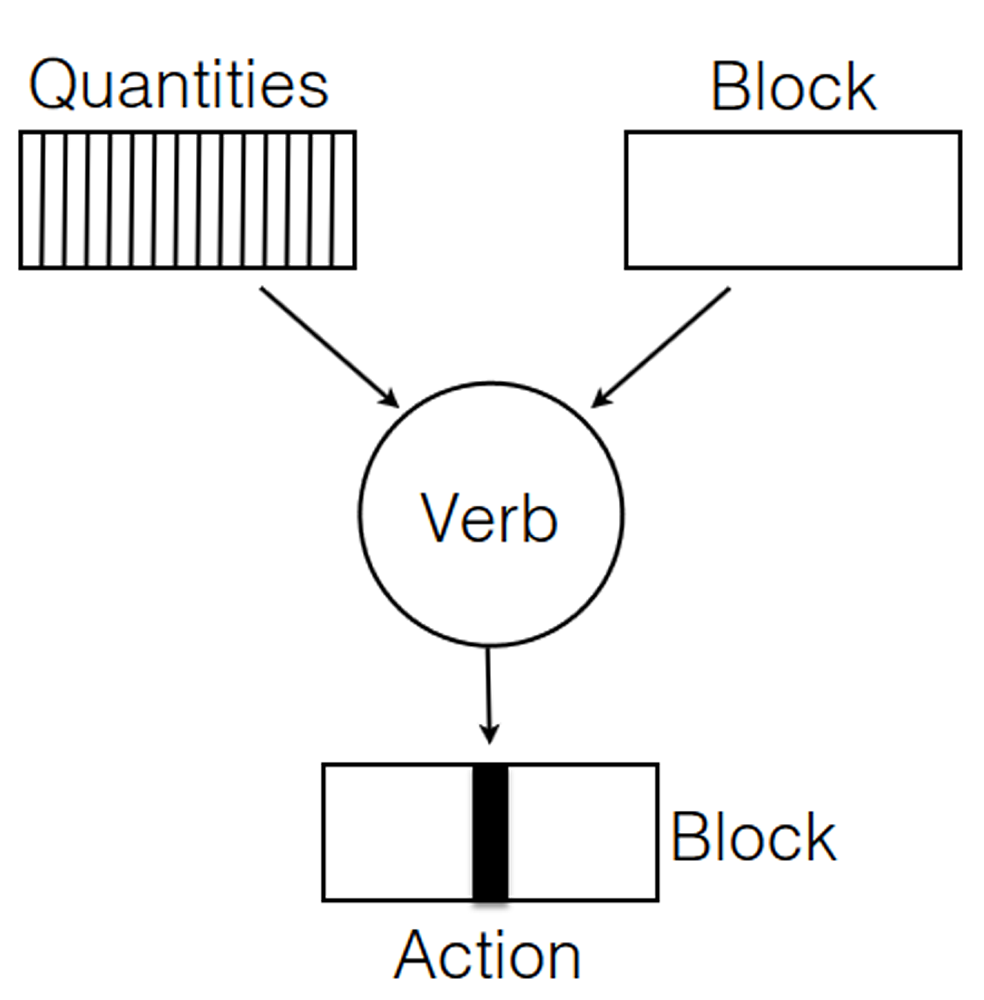}
\caption{Action of Verb}
\end{figure}
\begin{defn}
Let's call the mask set $\mu$, as an array of quantities $Q = (q_0,\ldots\,q_n)$ and an array of actions $A$, where each element of $A$ is an offset from the beginning of mask, as \first{Noun}
\end{defn}
\begin{defn}
Let's call the Noun set $\mathtt{N}$, Adjectives $(\mathtt{A}_0,\ldots,\mathtt{A}_k)$, and Verbs as \first{simple class}. Moreover, verb for simple class will be the ingress. Verb will use qualities and blocks as arguments.
\end{defn}
We note that an action may be employed, but not on every point of a layer.
\begin{defn}
Let's call a class without any verbs, or adjective, as an empty class. An empty class will contain only a noun.
\end{defn}
\begin{figure}[h]
\centering
\includegraphics[width=0.5\textwidth]{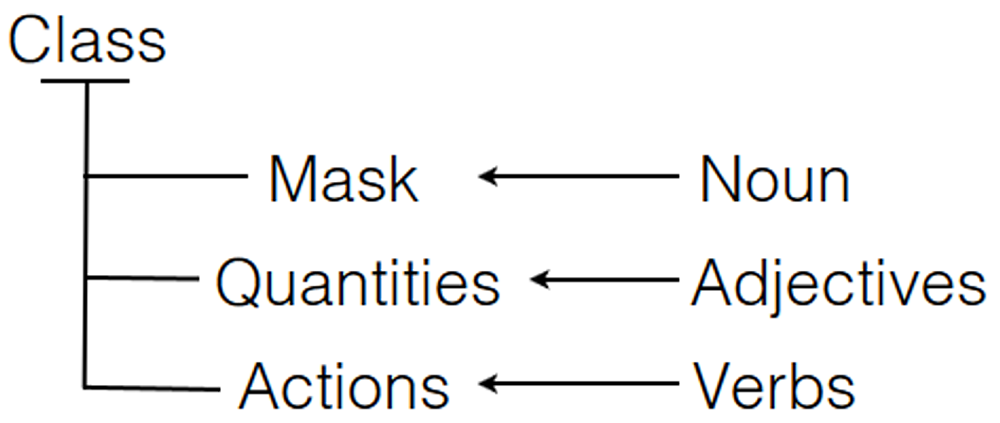}
\caption{Simple class structure}
\end{figure}
Under a simple class four operations can be performed:
\begin{description}
\item[Verb Addition] Add to class a new verb $\mathtt{V}'$ which equals $\mathtt{V}_l\xor\mathtt{V}_m$.
\item[Verb Multiplication] Add to class a new verb $\mathtt{V}'$ which equal $\mathtt{V}_l\band\mathtt{V}_m$.
\item[Adjective Addition] Add to class a new adjective $\mathtt{A}'$ which equals to $\mathtt{A}_l\xor\mathtt{A}_m$.
\item[Adjective Multiplication] Add to class a new adjective $\mathtt{A}'$ which equals to $\mathtt{A}_l\band\mathtt{A}_m$.
\end{description}
Let $\mu$ be the mask of a noun. And let $\mu_1^{\{i\}}$ be mask of another noun with the single difference in $i$ position. $\mu$ has constant $0$ or $1$ at $i$ position; $\mu_1^{\{i\}}$ has no constant at this address. Now it is possible for us to introduce two other operations:
\begin{description}
\item[Noun Specialization] The $\mu$ mask replaces to $\mu_1^{\{i\}}$ and the new adjective $\mathtt{A}_q$ adds to a class. $\mathtt{A}_q$ gets value from a layer by the address $i$.
\item[Noun Argumentation] The $\mu$ mask replaces to $\mu_1^{\{i\}}$ and the new verb $\mathtt{V}_q$ adds to a class. $\mathtt{V}_q$ sets value to a layer by the address $i$.
\end{description}
Due to finiteness of the quantity of verbs and adjectives in the class, it is possible to introduce the specific bijective function $\Phi$, such that it will display a class of operation to the layer.

If for a new adjective creation there is a unique operation, then the creation of the new verb has an internal ambiguity. It is because for a new verb we need to find a new point of action. In order to solve the ambiguity, we need to show one or more active adjectives, which are changed by the verb action point. The action point should be an argument of all active adjective and should not be an argument of all others. 
\begin{defn}
Let's set classes $\mathtt{C}$ and six operations between them. There is a fixed finite array of elements $(C_0,\ldots,C_N)$, which are called as basis. Now we can introduce the subset of classes $\tilde{\mathtt{C}}\subset{}\mathtt{C}$, which can be created from  the basis, by a finite count of operations. And we can introduce bijection $\Phi$ from $\tilde{\mathtt{C}}$ to the natural number set $\mathbb{N}$. Let's call it as \first{overclass mapping}.
\end{defn}
The specific form of $\Phi$ mapping is not important for the further consideration, note that it is invariant during the whole life. 
\section{Layers Working}
\label{sec:layers_working}
The mind system is made of stack and common objective memory. First, let's describe layers and their interactions. Null or signal layer is responsible for working with sensory and motor neurons. They excite and inhibit under the influence of the external environment. The current job does not put a purpose to solve the problem of response, so we will not consider the generation of egress signals.

Let's consider layers by induction. There are three layers, $i+1$, $i+2$ and $i+3$. For brevity let's call them as 1st, 2nd and 3rd, and remembering that they are not absolute numbers of layers.

On the 1st layer, there seems to be an excitement of elements. Thereafter, regularities are detected on a layer by overlaying noun's masks. Let all possible nouns be detected and the whole layer will be covered by blocks. It is not necessary to block all points of layer, some points can be free from blocking. 

After this, on the  2nd layer, masks are forcibly excited of the same nouns in the same order, but without duplicated elements and gaps. \ie{} if on the 1st layer nouns created in a sequence of $(A,B,B,B,C,C,D,A,C)$, then at the 2nd layer there will be a sequence of $(A,B,C,D,A,C)$.

On the next stage, the excitement state will be transmitted to the 3rd layer, also by detecting the noun's masks. Let the sequence $(A,B,\ldots,A,C)$  be responsible for the noun $E$, and this sequence be called  \first{sentence} of $E$. It is the simplified version, in any case the nouns which will be important of 2nd layer, are those which were covered in the ask of the 3rd.  
\begin{reqr}
Masks, which are forcibly excited on the layer, will not be detected on it.
\end{reqr}
\begin{figure}[h]
\centering
\includegraphics[width=0.5\textwidth]{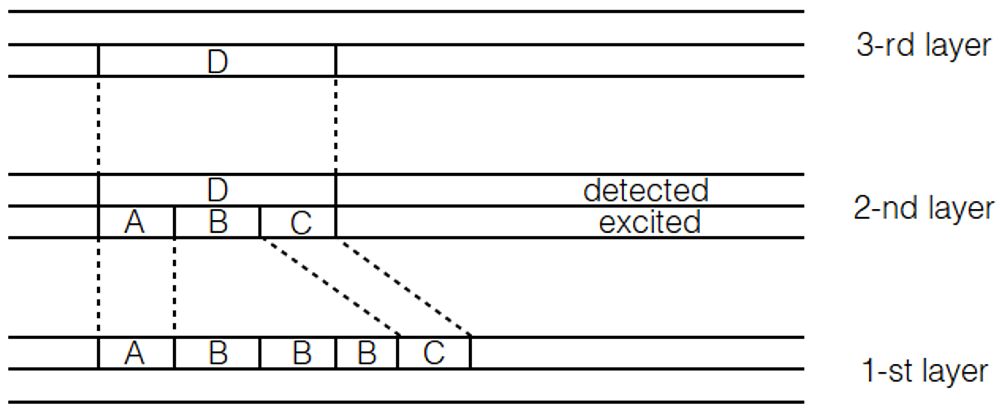}
\caption{Object detection}
\end{figure}
\begin{defn}
Let's call the simple class example which has specific qualities and actions as simple \first{object}. Every simple object is linked to the simple class, on which it acts. In the current examination objects were created on 1-st, 2-nd and 3-rd layers.
\end{defn}
\begin{defn}
Let's call the set of the simple classes and simple objects, which are denoted by the classes, as 0-level \first{context}. Let's call the pair of simple class and 0-level contexts as \first{complex class} 1-level. 
\end{defn}
Context is made on the 3rd level, made in the 1st and 2nd levels, in which further changes are contained. Now the 2nd layer controls the classes of context and 1st layer controls the objects.

Let's consider the behavior of objects of the 1st layer. They are in one context of 3rd layer, so they are connected together. In the beginning, they load their qualities from appropriate blocks of the 1st layer. In pseudocode:
\begin{algorithmic}
\ForAll{object $o$ in context}
        \ForAll{adjective $a$ in $o$}
                \State let $q_a$ is quality of $a$
                \State let $B(o)$ is block of $o$
                \State $q_a\gets a(B(o))$
        \EndFor
\EndFor
\end{algorithmic}
Then placement of verbs happens. Let the verb $f$ acts on the class $D$, and its quantities $(q_0,\ldots,q_n)$ be its arguments. Let $B(d)$ be the block of the object $b$. Then we should find classes with required quantities in context. Let class $C_b$ contain such quantities. Then objects of the class $C_b$ act on each object $d$ of class $D$: 
$$
d\to{}d'=f(B(d),b_i) \text{ for each }i\in(0,\ldots,k)
$$
In pseudocode:
\begin{algorithmic}
\ForAll{object $d$ in context}
        \ForAll{verb $v$ in $d$}
                \State let $a_v$ is action of $v$
                \State let $B(d)$ is block of $d$
                \State{let qualities $(q_0,\ldots,q_k)$ are linked with $v$}
                \ForAll{object $b$ in context}
                        \State{let $C_b$ is a class of $b$}
                        \If{$C_b$ has qualities $(q_0,\ldots,q_k)$ }
                                \State $a_v\gets v(B(d),b)$
                        \EndIf
                \EndFor
        \EndFor
\EndFor
\end{algorithmic}
It is very important to note that if verbs change the state of the block, instead of adjectives which change the objects, then it affects the following verb's actions. 
\section{Context Working}
\label{sec:context_working}
Let's introduce a series of definitions by induction:
\begin{defn}
Let's call a pair from the set $(N-1)$-level a complex class and the set $(N-1)$-level a complex objects as $N$-level \first{context}. Let's call the pair from a simple class and set of pairs of $(N-1)$-level complex objects and $(N-1)$-level contexts as $N$-level \first{complex class}.
\end{defn}
\begin{figure}[h]
\centering
\includegraphics[width=0.7\textwidth]{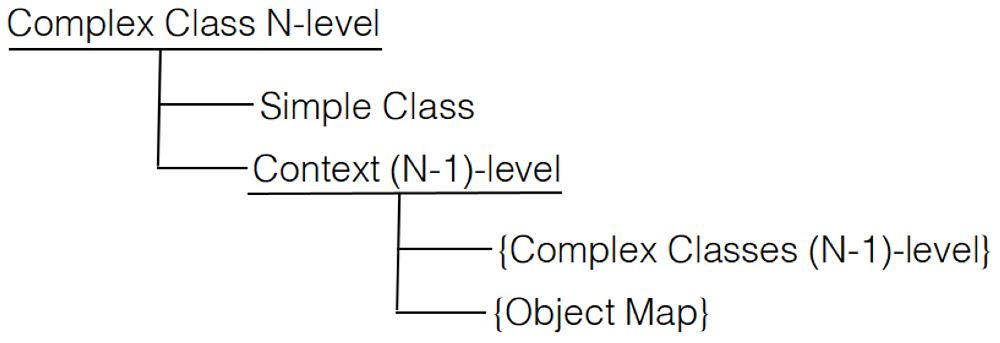}
\caption{Complex Class}
\end{figure}
In previous section there was a mind stack which contained 3 layers. Later we will consider a mind stack which will contain 6 layers. It this case verbs actions can be applied on 4 lower layers. Let us understand what will happen to them sequentially. 

Suppose that required masks were identified at the lowest layer, and after that masks were found on all layers, including the 6th layer. Now let's consider what happens at the 2nd layer. Because contexts are found on the 4th layer, the 2nd one has objects with adjectives and verbs. So excitation will spread onto the 2nd layer. The points on the 2nd layer are responsible for classes on the 1st layer. And due to existence function $\Phi$, which connects data with class operations, modified classes will act on the 1st layer rather than original ones. 

Since basic classes were in a global context, they were chosen from common memory of the system. So the modified classes on 2nd layer are already local ones for the their contexts. The local classes are placed into context and completely replace the lower layers for global classes with the same masks. Further operations on local classes only fulfill this context.

The next stage begins when the 5th layer happens. After this, it is possible to modify objects on the 3rd layer. So what happens when object with attached context are changed? There are 2 options: current context is modified or new empty context is created. Mentioned above, every object has qualities and actions. So it is possible to have two situation, each with its own named:  
\begin{description}
\item[Static] New context will be created if any quality on object is modified. Object with old attached context be placed into memory and will be available when specific qualities combination is met. 
\item[Dynamic] New context will be created if any action on object is modified. Object with old attached context be placed into memory and will be available when specific actions combination is met. 
\end{description}
In every mind system only one version of the system happens. It does not change during the lifetime~\cite{MayerBriggs}. Such a case is called \first{option}, this option is between static and dynamic. Let's call it \first{context option}.

In the layer system, described in the previous section, the upper layer creates classes on the lower one. In context memory the reverse process takes place, where upper class is modified by a set of classes from the lower level. For that there are so-called \first{spontaneous} class operations. Every mind system has 3 spontaneous operations, one for nouns, one for verbs, and one for adjectives. Let's introduce appropriate options:

For nouns:
\begin{description}
\item[Ratio] Noun Specialization operation is spontaneous.
\item[Irratio] Noun Argumentation operation is spontaneous.
\end{description}
For verbs:
\begin{description}
\item[Logic] Operation XOR for verbs is spontaneous.
\item[Ethics] Operation AND for verbs is spontaneous.
\end{description}
For adjectives:
\begin{description}
\item[Intuition] Operation XOR for adjectives is spontaneous.
\item[Sensorics] Operation AND for adjective is spontaneous.
\end{description}

Now it is possible to show how spontaneous operations work. First, we introduce some notations. Let's represent an object by the symbol as $O$ and classes by $C$. Upper index of classes and objects will represent a level. Apostrophe in classes $C'$ will represent what class is modified with the respect to the global class of $C$.

The 3rd layer has an object $O^3_1$ with a class ${C^3_1}'$. The object $O^3_1$ has context within which there are objects $\{O^2_i\}$ and the classes $\{{C_0^2}',{C_1^2}',\ldots,{C_N^2}'\}$. The task is to modify the class ${C^3_1}'$ and the $\{{C_0^2}',{C_1^2}',\ldots,{C_N^2}'\}$ so that the methods (verbs and adjectives) of ${C^3_1}''$ could create lower level classes, and to modify lower level classes, so they could create objects $\{O^2_i\}$. So the objects have to be immutable and classes in context could be modified. Later we will describe the algorithm step by step. But it should be noted that the algorithm creates not the only result, but many available results, and only the final step will choose the most appropriate one.

\textbf{Algorithm Begin}

\textbf{Step 1} Described previously, there is a bijection $\Phi$ between natural numbers and class, based on a global basis. Now consider the inverse transformation of $\Phi^{-1}$, which presents the class as ones and zeros. Let call it as binary representation of class. We will give the table a conventional example of the changes taking place.

\begin{center}
\begin{tabular}{ll}
${C^2_0}'$ & $10101010\ldots10$ \\
${C^2_1}'$ & $11101010\ldots11$ \\
$\ldots$ & $\ldots$ \\
${C^2_N}'$ & $10101110\ldots00$ \\
\end{tabular}
\end{center}

\textbf{Step 2} Let's move to the class ${C_1^3}'$. It has some adjectives, each of which take on a binary representation, and as a result gives meaning of quality. It is necessary to create new adjectives via adjective spontaneous operation, which will distinguish all different binary parts through ${C_1^3}'$ qualities.
\begin{center}
\begin{tabular}{llll}
                        &                  & $q_0, q_1,\ldots$        & \\ 
${C^2_0}''$ & $0101$ & $001\ldots10$ & $10$ \\
${C^2_1}''$ & $0101$ & $101\ldots11$ & $10$ \\
$\ldots$ & $\ldots$ \\
${C^2_N}''$ & $0101$ & $010\ldots00$ & $10$ \\
\end{tabular}
\end{center}

\textbf{Step 2.1} An additional step will be taken into action, if all possible algorithm results are considered invalid. In this step it is possible to introduce new adjectives which directly represent bits from layers to new qualities.

\textbf{Step 3} Before this step, let's remember that all classes in contexts were created from global contexts, by a series of verbs transformations. It is very important to note that verbs from ${C^3_1}'$ were used sequentially, when action from one was argument for another. Now it is necessary to do so what lower class will be created faster.

Let there be two verbs $V_1\to A_1$ and $V_2\to A_2$, which create actions  $A_1$ and $A_2$. Also $A_1$ is an argument for $V_2$. Therefore the verb $V_1\circ V_2\to A_2$ leads to the required value of $A_2$. A subtask of this step is to find a way, how it is possible, by using spontaneous verb operation, to create new verbs without intersections, by its arguments, and the ability to create classes for its contexts or parts of them.

\textbf{Step 3.1} is also an additional step for invalid results. Let there be context attached to $O_1^4$ and it be contained in the context of $O_1^3$. And let this context detect sentences where the quality from $O_2^3$ has always the same value with the bit $x$ from the block of $O_1^3$. Then in the mind, after a few tries, the system will create a verb, which connects quality to action over bit $x$.

\textbf{Step 4} All previous steps gave some set of possible realizations of class ${C_1^3}^{?}$. But the final version will be chosen after a uniqueness and an accuracy check. The memory system waits when the object $O_1^3$ will be loaded to layer. The upper layer will contain the sentence with $O_1^3$. The lower layer will contain a set of sentences with the set of that class. If all lower classes are in proposed realization, then it will be chosen as ${C_1^3}''$ instead of ${C_1^3}'$.    

\textbf{Algorithm End}

The process is simplification of memory system, it reduces the whole classes count and creates new verbs and adjectives. It called as \first{abstraction}. It is important to note that division into spontaneous and non-spontaneous operations required for solving problems with infinite looping. If count of nouns, verbs and adjective is finite then system will have finite count of there derivatives by spontaneous operations. The different combination of context creation and spontaneous operations options give 16 variants of behavior of the entire system. 

The main packing algorithm was described above in detail. However there is an additional process that named as \first{detalisation}. During detalisation steps 2, 2.1, 3, 3.1 are replaced with the following one:

\textbf{Algorithm End}

\textbf{Step 2 bis} Binary representation of all classes in context is modified by using noun spontaneous operation with condition that modified classes should describe all of its objects $\{O_0^2,\ldots,O_M^2\}$ in same context. The aim is to find same parts in all classes of the binary representations. All later steps will only be about the differences of binary classes representations. 

\begin{center}
\begin{tabular}{llll}
${C^2_0}''$ & $0101$ & $001\ldots10$ & $10$ \\
${C^2_1}''$ & $0101$ & $101\ldots11$ & $10$ \\
$\ldots$ & $\ldots$ \\
${C^2_N}''$ & $0101$ & $010\ldots00$ & $10$ \\
\end{tabular}
\end{center}

\textbf{Algorithm End}

\section{Structure of Input-Output System}
\label{sec:structure_of_input_output_system}
Let's analyze/look at the input-output structure in detail. As we know, there is a vast amount of signals from muscles and receptors; the overwhelming flow can hardly be directly processed by human consciousness. For primary processing of the flow there are \term{classical conditionings}. They can be expressed in the following chain:

\begin{figure}[h]
\centering
\includegraphics[width=0.7\textwidth]{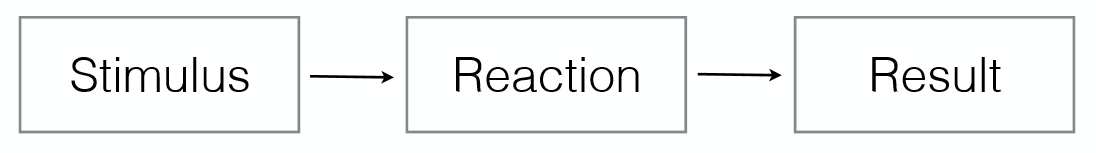}
\caption{Classical conditioning}
\end{figure}
The system of all classical conditionings is \term{the dynamic stereotype}. The dynamic stereotype plays multiple roles:
\begin{itemize}
\item It protects the consciousness from the huge flow of information from all the neurons in the body. 
\item It takes care of actions where comprehension is not necessary, including simple reactions like stepping, saying simple sounds, or "hit or run" reactions.
\item As a result of its work, the consciousness, instead of receiving certain signals, receives whole images, which are recognized at the level of the dynamic stereotype.
\end{itemize}
Let's try to construct a classical conditioning from the suggested model. For this purpose a special class where all verbs are connected to its qualities will be suitable. We will call it \first{the conditioning class}. A conditioning class is \term{a deterministic finite automaton}. A conditioning class should not only react to stimuli, but it should send signals to the consciousness. Therefore a conditioning class is a self-acting class, which processes signals from receptors and reports about some combinations of the signals to the consciousness. 

Therefore, all input-output can be divided into two segments: input segment and output segment. The input segment receives all signals from already working conditioning classes; the output segment is a space for generation of conditioning classes, which will be loaded to dynamic stereotype. Also it will be possible to determine on the output segment whether a class is suitable for the dynamic stereotype or not. If a newly generated class has a verb which requires any external qualities, then the class will not be conditioning and will not upload to dynamic stereotype. The algorithm of a working conditioning class is the following:

\textbf{Algorithm Begin}

\textbf{Step 1} A conditioning class $R$ uploads to dynamic stereotype. 

\textbf{Step 2} $R$ receives access to motor and sensor neurons, also $R$ receives a private area in signal layer for its output. 

\textbf{Step 3} $R$ detects images from sensor neurons, influences motor ones, outputs data to signal layer of the stack.

\textbf{Step 4.1} If $R$ detects all images successfully, after some time it will be unloaded from dynamic stereotype by a next command from the stack.

\textbf{Step 4.2} If $R$ does not detect a signal from sensor neurons (\ie{} its mask failed to apply), it will be forcibly unloaded, and will be interpreted by input of signal layer as \first{a failed} one. Processing of a failed class will be described in the next section in detail. 

\textbf{Algorithm End}

The main function of the dynamic stereotype is automated processing outside the consciousness. However, there is one more capability of it. After all, a human cannot only pronounce sounds, but he can think about how he pronounces sounds. This second option is called \term{internal speech}. 

Let us examine internal speech in more detail. A human can play out actions in his head, cause sensations, conduct mental operations, along with an internal monologue. These mental actions are absolutely similar to ordinary ones; they can cause the same recoil, but do not involve any muscles. Internal speech is that of a conditioning class which receives all signals from receptors, but all its out-going actions are connected to the input of the signal layer, not to muscles.

After examining this ability, we have come to the second feature of the output -- \first{speech direction}. It turns out that the output segment of the signal layer consists of a set of pairs -- conditioning class and speech direction. Speech direction serves for determining where a conditioning class will be uploaded for work. Speech direction has two options: \first{the internal speech} and \first{the external speech}. In the case of external speech a conditioning class will directly work with signals from muscles and receptors. 

Now let us remember that a mentally healthy human being has two qualities: 
\begin{itemize}
\item First of all, a human has only a single internal voice;
\item Second of all, the human is able to distinguish an internal voice and his fantasies from the reality.
\end{itemize}
Therefore, the internal speech has two important qualities: 
\begin{reqr}
The internal speech can process only a single class.
\end{reqr}
\begin{reqr}
The Output of internal speech is directed to a specially selected area of the signal layer, that is not intersected with signals from external speech.
\end{reqr}

\begin{figure}[h]
\centering
\includegraphics[width=0.7\textwidth]{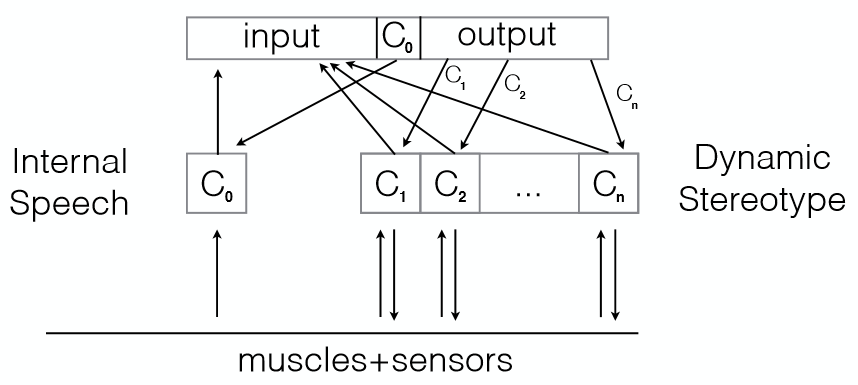}
\caption{Input and output}
\end{figure}
Internal speech performs a number of important functions which extends the model:
\begin{itemize}
\item Internal speech permits to mentally conduct operations that are not spontaneous for the memory. For example, it is possible to create a conditioning class that conducts an operation of verb addition XOR for an ethical organization of the memory. Using this class does not substitute for the operation, but a human will have the possibility to remember some rules that need to be done in certain circumstances,which will give the right result.
\item Internal speech also serves for control of the memory and the stack. By continuously uploading some information to the signal layer, the human is able to process the information as detailed as possible. 
\end{itemize}

\section{Decisions}
\label{sec:decisions}

The memory model has the algorithm of context packing that is able to create more complex classes from a set of lower ones. In most cases the algorithm is multivalued, \ie{} it can give multiple results. It has two types of multivaluency: 
\begin{itemize}
\item Firstly, multivaluency of operations. This is when it is possible to create multiple combinations of nouns, verbs and adjectives that are able to create all lower classes in some circumstances. 
\item Secondly, multivaluency of data. It is when its impossible to determine a point of action by the suggested adjectives.
\end{itemize}
In order to describe the multivaluency, let us suppose that using the packing algorithm creates all possible classes. We denote the whole memory tree in an initial state as $T$. In addition, the $\{p_{i}T, i=1,2,\ldots,N\}$ set will contain all possible packed trees with all possible packed classes. We will have marked with $p$ any changed tree $T$, and index $i$ enumerates all possible variants. In particular, the packing algorithm acts as the function
$$
T\to\{p_{i}T\}
$$
We will call any change \first{a patch}. This term presumes that packing generates $N$ different patches. 

Every patch is described by packed class $\tilde{C}$ and its context. The $\tilde{C}$ class has more nouns and adjectives than the original $C$, but class's context decreases. Moreover, the creation of a set of $\{\tilde{C}_i\}$ classes with difference in an action point, is possible. We mark it as a possibility, but the specific implementation of multivaluency will not be important later. It’s Important is that there is a finite set of all values in multivaluency.

\begin{figure}[h]
\centering
\includegraphics[width=0.7\textwidth]{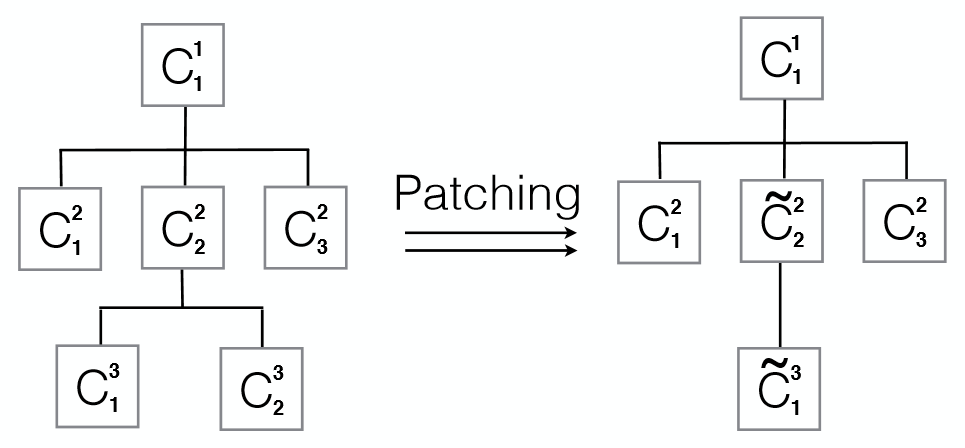}
\caption{Patching}
\end{figure}
However, packing can be applied not only to the original $T$ tree, but to any earlier packed tree. Let $p_{i}T$ be packed again, and $\tilde{C}$, changed at first stage, will be changed one more time. The result of the packing will be a set of memory trees $\{p^{1}_{j}p_{i}T, i=1,2,\ldots,M\}$. It will be the second level patch. We will denote $n$-level patch as a $\pi^{n}T$.

Consequently, because all patches has all different classes that are don’t intersect with each other, all possible patches $\pi{}T$ will form a structure like a tree. We will call the tree \first{the decision tree}. The original state of the memory without any modification is a root of the decision tree; leaves of the decision tree are final states of the memory with all possible modifications, let us denote them as $\hat{\pi{}}T$.
\begin{figure}[h]
\centering
\includegraphics[width=0.7\textwidth]{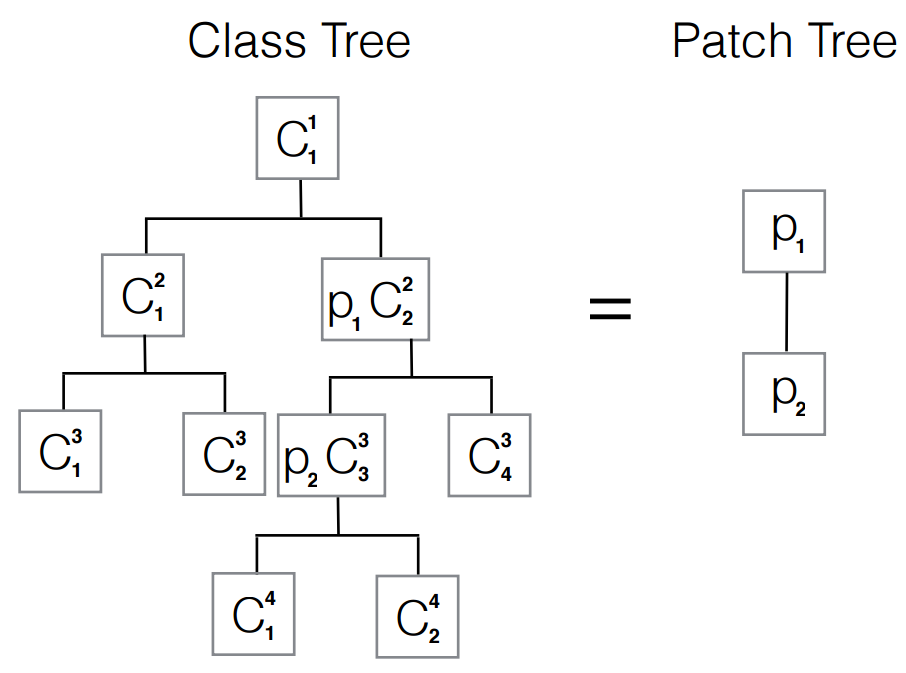}
\caption{Tree of patches}
\end{figure}

Let us examine the making of a decision. Let’s say, $\sigma$ was the initial state of the signal layer's input. We may say that the $\sigma$ applying to the memory tree $T$ led to the decision tree generation, because $\sigma$ exactly had additional data, that helped to simplify the original state of the memory $T$. It proceeded by the following algorithm:

\textbf{Algorithm Begin}

\textbf{Before Steps} The stack is clear, but the memory has a state as $T$.

\textbf{Step 1} Data is coming to the input of the signal layer. The signal layer has a $\sigma$ state; all levels of the stack are filling with decoded data.

\textbf{Step 2} During the process of filling upper layers', the memorized classes are addressed, that are modified according by the information from the $\sigma$. The memory tree is now set to $\tilde{T}$. 

\textbf{Step 3} The Modified state of the memory tree $\tilde{T}$ is packed by spontaneous operations. As a result of packing, there is a set of possible memory trees $\{\hat{\pi_i}T, i=1,\ldots,N\}$.

\textbf{Step 4} Each of the finished $\hat{\pi_i}T$ memory trees according to the signal, generates the segment’s condition, which follows the regular algorithm of creating a response. You can say that the $\sigma$ signal creates more than one response, but instead a host of possible $\{\kappa_i, i=1,\ldots,N\}$ responses.

\textbf{Step 5} The algorithm is branched here.

\textbf{Step 5.1} If $N = 1$, \ie{}, meaning the memory packed in the only possible way, then the memory will be fixed. A fixing of the memory is replacing the original $T$ state by the changed one $\pi{}T$; and the $\kappa$ response is generated on the output of the signal layer. The algorithm is finished here.

\textbf{Step 5.2} If $N > 1$, there is a set of possible options. Because it is impossible to generate multiple responses at one time, the following happens. All possible $\{\kappa_i\}$ responses from all possible packed trees will come to the signal layer again, but not to the usual output. They all together will come to the special zone of input; and the memory will be $T$ again. Let us call it the special zone, which is responsible for resolution of multivaluency, as an \first{Ego}. The algorithm is recursively iterated here with the new data; the return from the iteration will be to the next steps, denoted as post-steps.

\textbf{Step 5.2 post 1} If the result of the step 5.2 is the $\kappa_l$ response, then the final state of the memory will be $\pi_l{}T$.

\textbf{Step 5.2 post 2} In case none of suggested responses were chosen at step 5.2, but during the data processing the new $\pi^{(1)}T$ patch was created and the patch gave the only one response; then the patch will be chosen as a final state of the memory.

\textbf{Step 5.2 post 3} If there is no single response, all data is dropped and a response is not generated. 

\textbf{Algorithm End}
\begin{figure}[h]
\centering
\includegraphics[width=0.7\textwidth]{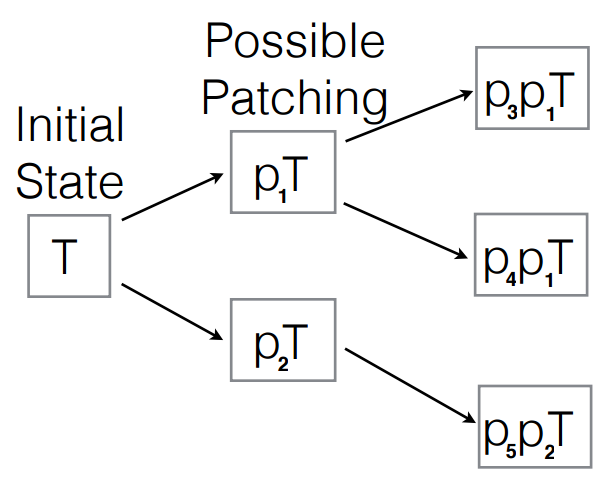}
\caption{All possible versions of memory tree after patching}
\end{figure}

\section{Model Training}
\label{sec:model_training}

Taking a more practical approach to the task  will allow us  to create a model. Let us look into a class structure that should contain all data and operations, where  operations should be able to convert to data and vice versa. . It is shown in the following, more precise, picture:

\begin{figure}[h]
\centering
\includegraphics[width=0.7\textwidth]{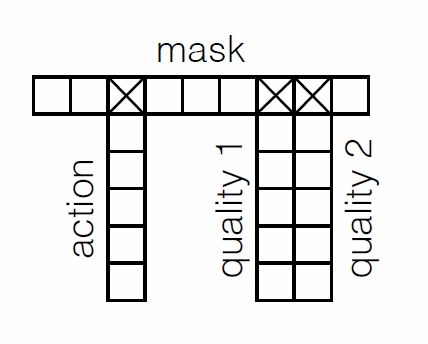}
\caption{Class in 2D}
\end{figure}

The main part of this section will be devoted to abstraction packing, \ie{} spontaneous operation will be conducted on verbs and adjectives. However the whole process is correct for detalisation packing also. 

One of the basic model requirements is the possibility to determine the same actions and qualities in different classes. The simplest way to solve this problem is to attach a global noun with its own mask to each action or quality. Thus, we will be able to reach uniform interaction by connecting same masks of different classes’ objects. It could be described by the scheme:  

\begin{figure}[h]
\centering
\includegraphics[width=0.7\textwidth]{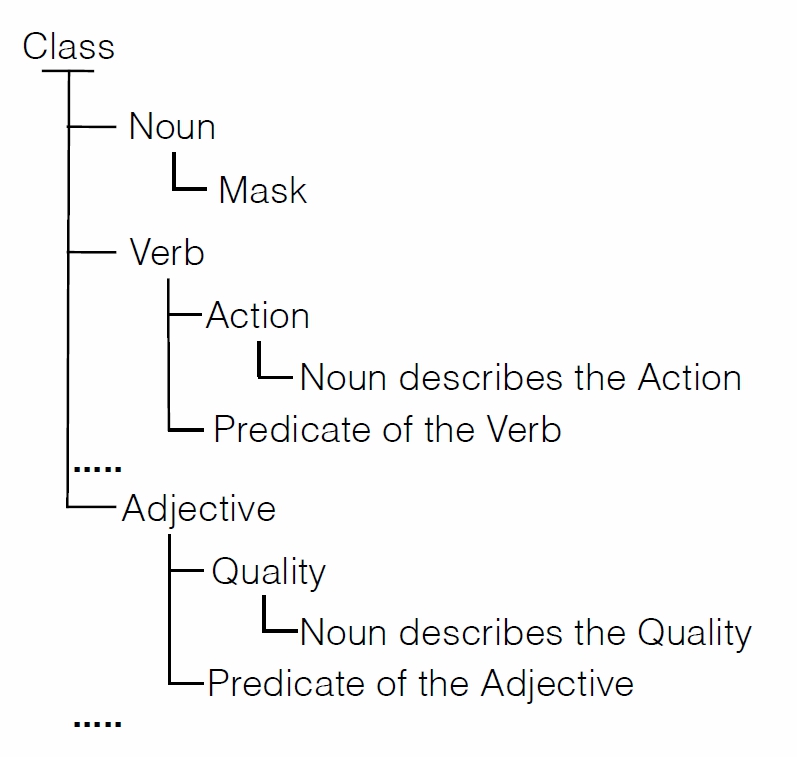}
\caption{Scheme of Class}
\end{figure}

Therefore, we should formalize these two processes:

First, the $Op$ operation on the following predicates: 
$$
NewPredicate = OldPredicate (AND/XOR) OtherPredicate 
$$
where $OtherPredicate$ is described by the $mu$ mask of its noun to change predicates of current classes.

Second, creation of a new mask for new qualities and actions necessary for  extension of  current classes.

Let us remember that a class can be in three states, when it is child on the stack, when it is parent on the stack and when it is stored in the memory. Parent state is the simplest one, at this moment class's object reads qualities from the lower layer and applies actions to them. It was described above in detail. 

The child state is more difficult than the parent one, it implies that class itself can change. Operations can be applied to the qualities of verbs and adjectives; new actions and qualities can be created. In addition, the child state should be responsive to the static-dynamic option to get access to different variants of nested contexts in the memory. The easiest way to express it would the following: 

\begin{enumerate}
\item Mask of the noun
\item Adjectives in binary format
\item Verbs in binary format
\end{enumerate}

What is  important here, is the ingress action points. Action points are parameters that can be used to change classes by writing values into them. Every action point has two parts, operational bit and mask of the argument. If operational bit = 1,  then  $AND$. If operational bit =  0, then $XOR$.

The third state is class's predicates serialization. Adjective's serialization can be carried out as follows:

\begin{enumerate}
\item Arguments' offsets
\item Operations in a binary format
\item Mask of the noun that describes the quality
\end{enumerate}

And verb's serialization as follows:

\begin{enumerate}
\item Masks of nouns that describe adjective
\item Operations in a binary format
\item Offset of the result action point
\end{enumerate}

For the ordering of binary operations we will use Zhegalkin's polynomial representation ~\cite{Zhegalkin}. Thus, all operations on predicates are stored as data.

However, the actual difficulty is not classes’ predicates storing but packing them in the memory. Its meaning is that a parent class should be converted to generator of one's nested classes. We have to order children's masks to allow operations from the parent class create all required operations. Moreover, there should be multiple variants of nested class's recombinations, if possible. It is possible because the nested classes have already been created from the global ones. However, it was a multiple-step process.

The most suitable option for development of such methods would be the genetic algorithm~\cite{Freitas}. The following parts are required to define the Genetic Algorithm:
\begin{enumerate}
\item spontaneous operations described earlier; they mix predicates of parent class for the best match to the test data;
\item fitness function that takes test data and runs it through modified parent class to determine the one that generates more nested than others;
\item test data that is various recombinations of masks of the nested classes.
\end{enumerate}

\begin{figure}[h]
\centering
\includegraphics[width=0.7\textwidth]{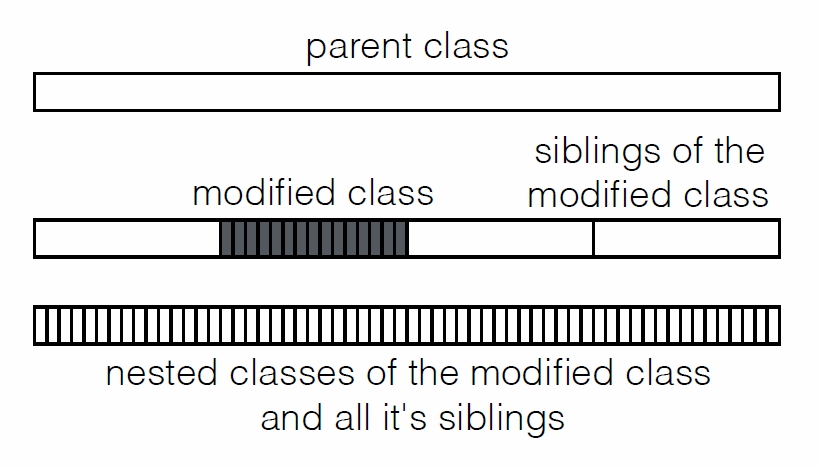}
\caption{Place of the modified class}
\end{figure}

The first two parts are already determined; now we need to sort the required test data. The main difficulty here is that fully functional performance requires  test data of all siblings of parent class and their nested and subnested classes. Therefore, generation of test data should be intellectual; simple recombination of the nested class would not work. 

It is suggested that system of neural network be used.  We will call it an Imaginator. Imaginator is a Generative Adversarial Networks~\cite{Goodfellow} and has two parts: a Generator and Discriminator.

The working loop of Generator $G$ is in the following algorithm:

\begin{algorithmic}
\State let $C$ is class
\State let $T$ is tree data for evolution testing
\ForAll{class $D$ in context of Grand-Parent of $C$}
		\State let $R_0$ are all children of $D$
		\State let $G(R_0)$ is recombination of $R_0$
		\State $G(R_0)$ MUST be covered by mask of $D$
		\State $G(R_0)$ MUST pass Discriminator
		\State put $G(R_0)$ to $T$
		\ForAll{unique class $P$ in $G(R_0)$}
				\State let $R_1$ are all children of $P$
				\State let $G(R_1)$ is recombination of $R_1$
				\State $G(R_1)$ MUST be covered by mask of $P$
				\State $G(R_1)$ MUST pass Discriminator
				\State put $G(R_0)$ to $T$
		\EndFor
\EndFor
\State $M(C)$ are all classes modified by Evolution
\State apply $T$ to $M(C)$ to get new modified contexts
\State choose the most appropriate one 
\end{algorithmic}

it is required that Generator $G(R)$ output corresponds with the mask of each parent class. However, it is not the only requirement. It is necessary that the Generator “could” pass the Discriminator. The Discriminator must ensure that Generator's output is realistic. For this purpose, the Discriminator learns on real data, specifically on combinations of nested classes that appear on the stack during work. The full learning process can be shown as follows:

\begin{enumerate}
\item Discriminator gets data for self-learning from the stack and trains Generator.
\item Generator makes test data for Genetic Algorithm.
\item Genetic Algorithm generates patches.
\item If patches are applied to the memory tree, the data from the patch are considered as good. 
\item If patches roll back, the data from the patch are considered as bad.
\end{enumerate}

As the result, we see a chain of the Imaginator and the Genetic Algorithm that creates patches for the memory tree. In general, the process is as follows: the Discriminator takes recombinations for learning from the stack and trains the Generator; The Generator creates test data for the Genetic Algorithm; the Genetic Algorithm creates changed parent classes that are packed to patches.

In addition, the Imaginator learns itself during the patch creation process. Apart from the standard learning process of the Generator and Discriminator, described above, there is an additional one. Let it be the set of patches $\{P_i\}$. If patches are applied to the memory tree, it would be a changed memory tree. This tree is then tested on the real data derived from the stack. If during the work flow the system reserves a positive motivation, then the changes are fixed. If the system does not receive a positive motivation, the changes are dropped. However, besides the positive motivation that reinforces the memory, there is pain. If the applied changes result in the pain syndrome, then $\{P_i\}$ patches are not only fixed, but also go to the  Discriminator’s input as negative; and the Disciminator learns to avoid them. Thus, the system is configured to prioritize pain avoidance over a motivational reinforcement.

Prospect theory~\cite{Kahneman} assumes that people intuitively analyze possible events in the wrong way. Low probabilities tend to be overvalued and the high ones tend to be undervalued. This effect should be observed in the model because the Generator creates recombinations of nested classes excluding the probability they occurrence in the real data. In addition, the overvaluation of negative possibilities should be observed because the pain effects are applied directly to the Discriminator.

For the sake of completeness, we should introduce one more neural network, the Detector. It is also GAN that contains two parts. The Detector's discriminator searches repeating sequences in the stack to create new masks. The Detector's generator creates new unique sequences, which would never appear on the stack, to mark new qualities and actions. The Detector’s generator learns with the help of its Discriminator. 

The Detector's working process includes work of genetic algorithm and Imaginator with the only difference in the packing. It uses detailisation packing instead of abstraction. 

To give the final definition of the system with Reinforced Learning, we have to describe the input data and its ways. The continuous working process should have two types of data: ingress data for routine operations and control data for approval/rejection of generated patches. 

Ingress data come from two sources:
\begin{enumerate}
\item Signals from motor and sensor neurons with information from the external world. This is quite a simple type of information and the system considers it as external. It is called \term{feeling}. 
\item Signals about total changes in the sets of applied and dropped patches. It has no precise knowledge about the information from the patches, but only indicators that show whether a patch was applied or dropped and in what order. This type of information is considered as internal. It is called \term{emotions}.
\end{enumerate}

The control data do not pass into the model directly. However, they affect the learning process. There are two types of control data:
\begin{enumerate}
\item Pleasure. If the system receives pleasure, it applies a patch and reinforces the Generator.
\item Pain. If the system receives pain, it drops a patch and reinforces the Discriminator.
\end{enumerate}
It should be noted that if ingress routine data come into the system continuously, the control data come sporadically. Most of the time the patches are created without any control data, the control data affect the final result only: whether a patch is going to be applied to the memory tree or not. 

\section{Dynamics}
\label{sec:dynamics}

All the previous sections described only one step in the whole system. However, the system should work continuously. Let us introduce the stack frequency. Obviously, the memory works with the same frequency. The memory packer and the Imaginator will work with proportional frequency. However, it is true not for all the system components. The Detector's frequency and operation sequence are not directly connected to the stack frequency. 

To conduct further analysis, let us remember that the brain consists not only of neurons, but it is also affected by hormones. Let us introduce two hormones: a happiness hormone and a sadness hormone. The hormones are excreted proportionally to applied or dropped patches. Let an applied patch $P_1$ excrete $H(P_1)$, and a dropped patch $P_2$ excrete$S(P_2)$. It is required to ensure a self-balancing model. If the process runs smoothly and all patches are applied one by one, the model needs a motivation for new data input. If the process fails and all patches are constantly dropped, the model needs a motivation to do something absolutely new that has never been done before.

Let us introduce two frequencies for the Detector. $\omega_{lG}$ is the learning frequency of the Detector's generator and $\omega_{lD}$ is the learning frequency of the Detector's discriminator:
$$
\omega_{lD} \appropto \sum_{time period}H(P_1)
$$
$$
\omega_{lG} \appropto \sum_{time period}S(P_1)
$$
where the sum is the total hormone accumulated over a certain period of time. Obviously, if the Detector's discriminator starts finding new stable sequences and creating new nouns, the number of errors should increase. 

The other components of the system's dynamic control are the excitation, depression and focusing. Consciousness does not directly participate in the processes, but affects them indirectly. The mechanism of the said influence is described below. Let us remind the reader of the global classes that we have introduced above. The global classes are classes that could be accessed from any private context. $\{G_i\}$ is a set of global classes. Let us match $p_i$ parameter to each global class and call it \first{priority of class}. We will call active priority at some point in time the sum of priorities of all global classes that are active at this point in time.
$$
P=\sum_{active}p_i
$$
Thus, the selection system can be introduced. If an active priority is above the average one, the system excites. If an active priority is below the average one, the system depresses. Similarly, we can now compare the priorities of two active contexts. 

We need to discuss the mechanisms of priority changing. We can define two mechanisms: external and internal. The external mechanism is quite a simple one. The priories of all active classes increase in the case of pleasure and decrease in the case of pain. This mechanism is used in animal training. An animal is fed when it performs the task successfully and punished when it fails to do so.

We have to say that up until this moment the model could be applied to any mammal. The only difference would be the quantity of available layers on the stack. A human should have 6 layers, monkey -- 5 layers and rat -- 2 layers. However, humans are different from animals in that they develop and preserve speech across generations. The speech does not depend on genetics, it is transmitted from mother to child and later keeps on developing during the whole life. Chimpanzees can be trained to produce some basic speech elements, but this speech is only preserved and used if monkeys contact with humans. 

The only explanation is that humans must have a unique motivation to speak and communicate. We should also note that the external mechanism of priority changing is closely connected with the survival needs. However, the human internal mechanism is likely connected with the need to procreate. The other fact to support it is that speech develops constantly and not only due to hunger or threat. Moreover, all human societies consider communication to be the most important aspect of courtship, whereas animals practically do not communicate before mating. We can even assume that human internal motivation prevails over the external one.

The internal priority change in a particular class occurs in the following independent cases. To increase: 
\begin{itemize}
\item If the Input-Output layer has been stably operating for a long time. In this case the situation is absolutely predictable. 
\item If a patch is applied. In this case the system receives new knowledge. 
\end{itemize}
To decrease:
\begin{itemize}
\item If the Input-Output layer is constantly updated. In this case the situation is absolutely unpredictable. 
\item If a patch dropped. In this case the system loses knowledge. 
\end{itemize}

The motivation is not limited to these two cases. It is necessary to understand how the system would behave if these actions had different directions. To do that, let us introduce the concept of gender. We should bear in mind that gender is a mental concept, unlike sex that is a biological one. Gender determines a person's self-identification at a very early age, even before puberty. Also, the existence of gender phenomenon is supported by transgender people who do not feel themselves in their bodies. That said, the body itself is healthy and its hormonal environment corresponds to the biological sex. 

For considering of motivation, let's return to self-action. In the beginning of this section, the details of available options the system acting to itself using excitation, depression and focusing was described. In addition, self-perception using feeling from mistakes being made and successes being reached. That gives wide spectrum of possible self-interaction. 

Now let's divide the whole Input-Output layer onto parts: $IO_A$ and $IO_B$, where the first part is external actions and feelings, and the second -- self-action and self-feelings being described previously. Thus in the common case, every function $f$ acting in the layer can be described as the following:
$$
f(IO_A, IO_B)\to (IO_A, IO_B)
$$
This function may have four special cases:
$$
\begin{aligned}
f_1(IO_A)\to (IO_A)\\
f_2(IO_B)\to (IO_B)\\
f_3(IO_A)\to (IO_B)\\
f_4(IO_B)\to (IO_A)
\end{aligned}
$$
Here we make the special hypothesis that World and Self stay same in common and are continuously interacting. In this case, those four cases are not independent because every change of World responses to change of Self and otherwise. Just their interaction is important. Note that this postulate is not a consequence of any previous, this is the requirement of correct definition of self-interaction and motivation. 

Then any function acting on the Input-Output layer can be divided to three unrelated parts, Self and Self, World and World, Self and World. Moreover Self and World interaction can be represented by two various ways:
\begin{description}
\item[Male type] World and Self interact with $f_3(IO_A)\to (IO_B)$ 
\item[Female type] World and Self interact with $f_4(IO_B)\to (IO_A)$
\end{description}

\section{Conclusion}
\label{sec:conclusion}
This model describes the process of understanding the human speech, but now it has the following problems:
\begin{itemize}
\item There is no strict experimental basis, which could show how different people process the information.
\item Practical implementation may be difficult because there is no clear understanding of the model's neural network structure.
\item The model has no hypotheses or ideas on how to start its implementation from scratch and the newborn reflexes role.
\end{itemize}

\section{Acknowledgments}
I thank Ilya Kruglov for editing of this manuscript. 

\bibliographystyle{alpha}
\bibliography{chistyakov15}

\newcommand{\etalchar}[1]{$^{#1}$}
\begin{thebibliography}{GPAM{\etalchar{+}}}

\bibitem[Fre02]{Freitas}
Alex~A. Freitas.
\newblock {\em Data Mining and Knowledge Discovery with Evolutionary
  Algorithms}.
\newblock Springer, 2002.

\bibitem[GPAM{\etalchar{+}}]{Goodfellow}
Ian~J. Goodfellow, Jean Pouget-Abadie, Mehdi Mirza, Bing Xu, David
  Warde-Farley, Sherjil Ozair, Aaron Courville, and Yoshua Bengio.
\newblock Generative adversarial networks.

\bibitem[Jun71]{Jung}
Carl~Gustav Jung.
\newblock {\em Psychological {T}ypes}.
\newblock Princeton University Press, 1971.

\bibitem[KT79]{Kahneman}
Daniel Kahneman and Amos Tversky.
\newblock Prospect theory: An analysis of decision under risk.
\newblock {\em Econometrica}, 47:263--291, March 1979.

\bibitem[MM95]{MayerBriggs}
Isabel~Briggs Myers and Peter~B. Myers.
\newblock {\em Gifts {D}iffering: {U}nderstanding {P}ersonality {T}ype}.
\newblock CPP, 1995.

\bibitem[Zhe27]{Zhegalkin}
Ivan~Ivanovich Zhegalkin.
\newblock On the technique of calculating propositions in symbolic logic.
\newblock {\em Matematicheskii Sbornik}, 43:9–28, 1927.

\end{thebibliography}
\end{document}